\documentclass[sigconf, nonacm]{acmart}
\usepackage{graphicx}
\usepackage{wrapfig}
\usepackage[symbol]{footmisc}
\usepackage{enumitem}
\usepackage{titlesec}
\usepackage{lipsum} 
\usepackage{caption}
\setcopyright{none}
\usepackage[draft,textsize=footnotesize,textwidth=15mm]{todonotes}

\AtBeginDocument{%
  \providecommand\BibTeX{{%
    \normalfont B\kern-0.5em{\scshape i\kern-0.25em b}\kern-0.8em\TeX}}}
\setcopyright{acmlicensed}
\copyrightyear{2024}
\acmYear{2024}
\acmDOI{XXXXXXX.XXXXXXX}
\title{AuditLLM: A Tool for Auditing Large Language Models Using Multiprobe Approach}
\begin{document}

\author{Maryam Amirizaniani}
\email{amaryam@uw.edu}
\affiliation{%
  \institution{University of Washington}
  \city{Seattle}
  \state{WA}
  \country{USA}
}
\author{Elias Martin}
\email{eamart34@uw.edu}
\affiliation{%
  \institution{University of Washington - Bothell}
  \city{Bothell}
  \state{WA}
  \country{USA}
}
\author{Tanya Roosta\textsuperscript{*}}
\email{tanyaroosta@gmail.com}
\affiliation{%
  \institution{UC Berkeley, Amazon}
  \city{Saratoga}
  \state{CA}
  \country{USA}
}
\author{Aman Chadha\textsuperscript{*}}
\email{hi@aman.ai}
\affiliation{%
  \institution{Stanford University, Amazon GenAI}
  \city{Palo Alto}
  \state{CA}
  \country{USA}
}
\author{Chirag Shah}
\email{chirags@uw.edu}
\affiliation{%
  \institution{University of Washington}
  \city{Seattle}
  \state{WA}
  \country{USA}
}

\thanks{\textsuperscript{*}Work does not relate to position at Amazon.}



\begin{abstract}
As Large Language Models (LLMs) are integrated into various sectors, ensuring their reliability and safety is crucial. This necessitates rigorous probing and auditing to maintain their effectiveness and trustworthiness in practical applications. Subjecting LLMs to varied iterations of a single query can unveil potential inconsistencies in their knowledge base or functional capacity. However, a tool for performing such audits with a easy to execute workflow, and low technical threshold is lacking. In this demo, we introduce ``AuditLLM,'' a novel tool designed to audit the performance of various LLMs in a methodical way. AuditLLM's primary function is to audit a given LLM by deploying multiple probes derived from a single question, thus detecting any inconsistencies in the model's comprehension or performance. A robust, reliable, and consistent LLM is expected to generate semantically similar responses to variably phrased versions of the same question. Building on this premise, AuditLLM generates easily interpretable results that reflect the LLM's consistency based on a single input question provided by the user. A certain level of inconsistency has been shown to be an indicator of potential bias, hallucinations, and other issues. One could then use the output of AuditLLM to further investigate issues with the aforementioned LLM. To facilitate demonstration and practical uses, AuditLLM offers two key modes: (1) Live mode which allows instant auditing of LLMs by analyzing responses to real-time queries; and (2) Batch mode which facilitates comprehensive LLM auditing by processing multiple queries at once for in-depth analysis. This tool is beneficial for both researchers and general users, as it enhances our understanding of LLMs' capabilities in generating responses, using a standardized auditing platform. 
\end{abstract}

\maketitle

\section{Introduction}

In the rapidly evolving landscape of artificial intelligence, Large Language Models (LLMs) have emerged as cornerstones of contemporary AI applications, driving innovations in natural language processing (NLP), automated content creation, and decision support systems~\cite{shi2023retrieval}. Their ability to understand, generate, and interact using human-like language has paved the way for unprecedented advancements across various domains. For example, LLMs have become increasingly integrated into critical sectors, such as healthcare~\cite{gu2021domain, thirunavukarasu2023large, yang2022large}, education~\cite{dai2023can, meyer2023chatgpt, kasneci2023chatgpt}, legal~\cite{liu2023ml} and financial services~\cite{zhang2023enhancing, huang2023finbert}. Given the mission critical nature of these applications, it is imperative to ensure the LLMs are reliable and robust. Recognizing this, the academic and tech communities have underscored the importance of rigorous auditing mechanisms to probe LLM performance. Previous research has laid a solid foundation for understanding the potential of LLM auditing, focusing on identifying biases~\cite{Motoki2023, Talat2022, Thakur2023}, hallucinations~\citep{chen2023hallucination, sadat2023delucionqa, yang2023new}, inconsistency \citep{Tam2023, Ye2023}, reliability \citep{Shen2023, Zhong2023study} and failures in model responses~\citep{rastogi2023supporting}. 

Building on the existing foundational work ~\citep{rastogi2023supporting}, we introduce a novel auditing tool, AuditLLM, which provides a general-purpose solution for auditing LLMs. A simple but effective approach for probing LLMs involves deploying multiple versions of the same query to identify any inconsistencies in the model's output. To implement this audit technique effectively, a mechanism for crafting these questions or probes is essential. 
Although humans are capable of formulating various ways to pose the same question for probing purposes, this approach is not scalable. Therefore, we adopt the technique described in~\cite{amirizaniani2024developing} to streamline probe generation using a different LLM.

\begin{figure*}[htbp]
    \centering
    \includegraphics[scale=0.52]{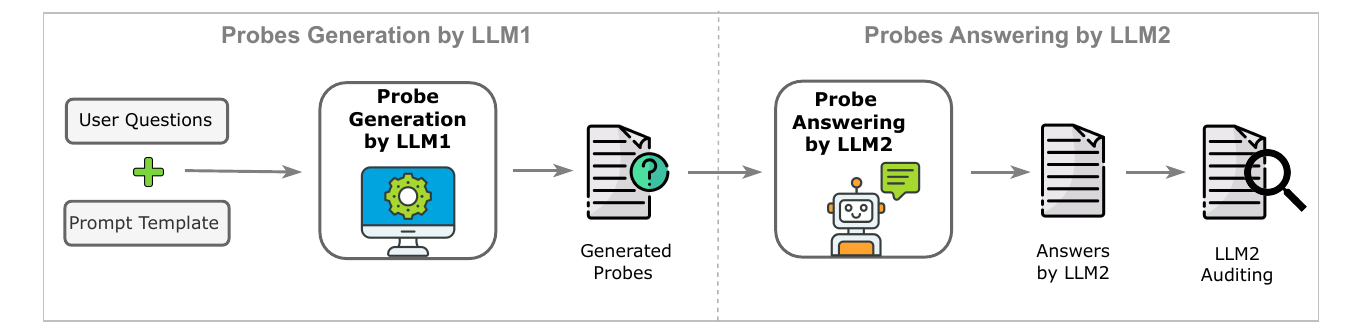}
    \caption{The AuditLLM system employs two LLMs: LLM1 generates probes and LLM2 provides responses to these probes.}
      \vspace{-0.5em}
    \label{fig:flow}
\end{figure*}

To facilitate this, AuditLLM offers two primary features: \textbf{(1) Live Mode:} This mode allows for real-time auditing, where a user's single inputted question is transformed into five distinct yet related prompts. The LLM's responses are then generated and assessed for semantic similarity, offering immediate insights into the model's potential inconsistencies; and \textbf{(2) Batch Mode:} Batch mode caters to a broader analysis by accepting multiple questions at once, evaluating the LLM's performance on a larger scale, and providing a semantic consistency score for each generated response. These innovative features of AuditLLM bolster the robustness, transparency, and ethical deployment of LLMs, thereby advancing the state-of-the-art in Artificial Intelligence.

The \textbf{motivation} for this research is to audit the inconsistencies in LLMs as reflected in question-and-answer tasks, so as to develop a sense for potential bias or hallucinations. The \textbf{novelty} of AuditLLM lies in its specialized approach to probing LLMs through the lens of inconsistency detection in the above-mentioned modes. Our \textbf{contributions} can be outlined as follows: 
\begin{enumerate}
    \item The live mode feature of AuditLLM allows for the immediate auditing of LLMs in real-time in an interactive way.
    \item The batch mode of AuditLLM supports an extensive evaluation of LLMs by enabling the submission of numerous questions simultaneously, conducting a detailed and comprehensive audit.
\end{enumerate}
In the remainder of this paper, we describe each mode, illustrating their capabilities to detect and highlight inconsistencies in the LLM generated responses.

\section{Related Work}

The growing application of LLMs in information retrieval (IR) highlights the need for a critical evaluation of their associated risks, a concern extensively recognized within the academic community~\cite{Bender2021, Blodgett2020}. Auditing LLMs, aimed at assessing specific attributes to mitigate these risks, has emerged as a crucial research domain. Studies, such as ~\citep{wang2023decodingtrust}, have undertaken thorough evaluations comparing models like GPT-4 to its predecessors in terms of toxicity, stereotype bias, and adversarial robustness. Nonetheless, existing auditing methods often lack generalization capabilities and require substantial human involvement. For instance, \citet{petroni2019language} and \citet{jiang2020know} employed fill-in-the-blank prompts to audit LLMs' knowledge of particular entities, but this method struggles with generalization when applied to generative tasks. This limitation stems from the dependency on the specific knowledge encoded within each model, restricting broader applicability across various LLMs. Additionally, initiatives like RealTocxicityPrompts\citep{gehman2020realtoxicityprompts} and ToxiGen\citep{hartvigsen2022toxigen} have developed benchmarks with toxic and non-toxic prompts to audit LLM output safety; however, these approaches also require considerable human effort, and the creation of new benchmarks for different domains. Similarly, ~\citep{blodgett-etal-2021-stereotyping} manually developed a codebook to categorize biases in benchmarks, and ~\citep{ganguli2022red} involved red team members in writing attack prompts to test LLM robustness, further illustrating the intensive human labor required in these methods.

Systematic auditing of LLMs has been explored by few studies. ~\citet{mokander2023auditing} introduced a three-tiered auditing pipeline to address the ethical and social challenges of LLMs, covering audits of model designers, pre-release LLM evaluations, and audits of applications using these models. Although conceptually robust, the framework provides limited practical implementation guidance. Further advancing this field, ~\citet{ribeiro-lundberg-2022-adaptive} introduced AdaTest, a collaborative process that combines human expertise and LLM capabilities to craft unit tests for auditing purposes. In this methodology, an LLM generates test suggestions based on existing examples within a given topic, which are then refined by users. Subsequent iterations of testing are organized using a predefined test tree, guiding the LLM in generating new test suggestions. However, this method is somewhat inflexible, depending solely on a single LLM for test generation and lacking a systematic approach to the creation of test questions.

Building on this, AdaTest++ ~\cite{rastogi2023supporting} enhances the original model by enabling users to specifically request test suggestions, thereby allowing for more targeted exploration of potential LLM errors and facilitating more purposeful guidance in subsequent testing loops. While AdaTest and AdaTest++ represent significant strides in creating a systematic auditing pipeline, they exhibit notable limitations: (1) the reliance on GPT-3.5 restricts the scope of LLMs auditable within these frameworks; (2) the random generation of test sets lacks a systematic approach and fails to meet specific auditing criteria; and (3) the methodologies are not suited for batch auditing of LLMs. Our proposed framework -- AuditLLM -- can effectively address these issues as described in the following section.

These efforts reflect advancements in LLM auditing for responsible use, however, none of these approaches have specifically targeted LLM inconsistencies. Our proposed framework, AuditLLM, has been designed specifically to address this gap in auditing LLMs. Given that inconsistencies could be harbingers for issues of bias and hallucination, focusing on auditing an LLM for its inconsistencies allows researchers and developers to be broader in their auditing efforts. 

\section{AuditLLM Tool Demonstration}

\begin{figure}
    \centering
    \includegraphics[width=0.35\textwidth]{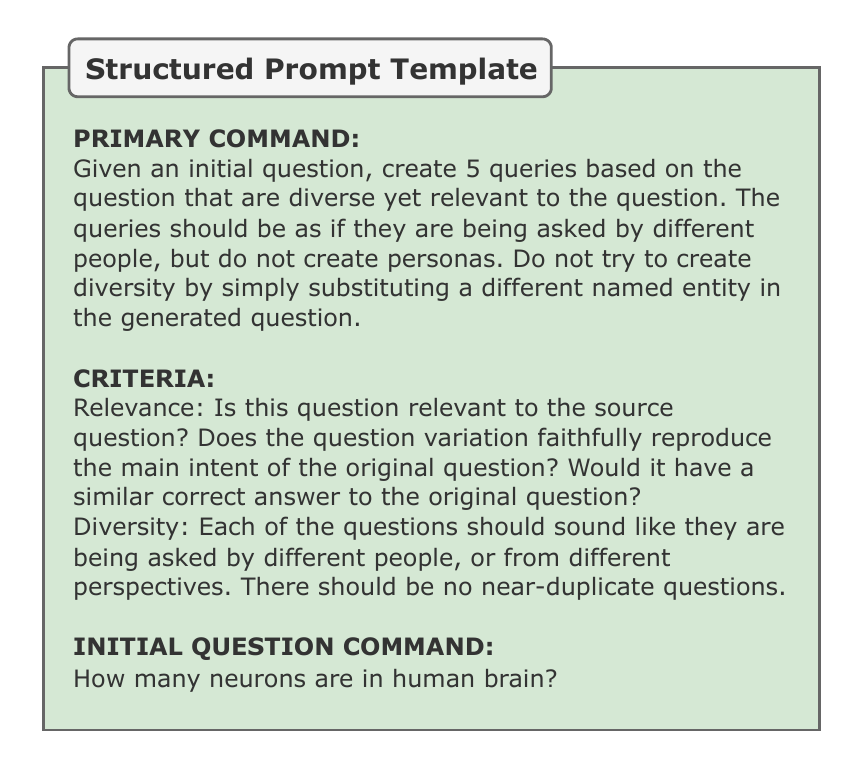}
    \caption{Prompt Template for LLM1 \cite{amirizaniani2024developing}.}
    \vspace{-10pt}
    \label{fig:template}
    \vspace{-5pt}
\end{figure}

AuditLLM\footnote{Implemented with the Gradio Library, this tool is available as a free, open-source project on Huggingface: \url{https://huggingface.co/spaces/Amirizaniani/AuditLLM}.}, serves as a systematic tool to reveal inconsistencies and conduct audits through the utilization of two LLMs: LLM1 and LLM2. As depicted in Figure~\ref{fig:flow}, LLM1 is tasked with the creation of probes based on the given prompt template, whereas LLM2 is responsible for generating responses to these probes. The prompt template for LLM1 is iteratively created using human-in-the-loop technique described by \citet{amirizaniani2024developing}, and shown in Figure~\ref{fig:template}. The code for AuditLLM is also available on HuggingFace\footnote{\url{https://huggingface.co/spaces/Amirizaniani/AuditLLM/blob/main/app.py}}.

AuditLLM introduces two distinct modes, live and batch, both designed to use multiple probes generated by LLM1 to detect discrepancies within the responses produced by LLM2. These modes delve into meaning comprehension of generated responses, enabling analysis of the model's consistency by AuditLLM. For a fair comparison, the \texttt{temperature} of LLM2 was set to 0.5 and \texttt{max\_length} was configured to 512. The functionalities of each mode are detailed below. A video illustrating the functionality is also hosted on YouTube\footnote{\url{https://www.youtube.com/watch?v=49UMQOd7Rm0}}. 

\subsection{Live Mode}

\begin{figure}[ht]
    \centering
    \includegraphics[width=0.40\textwidth]{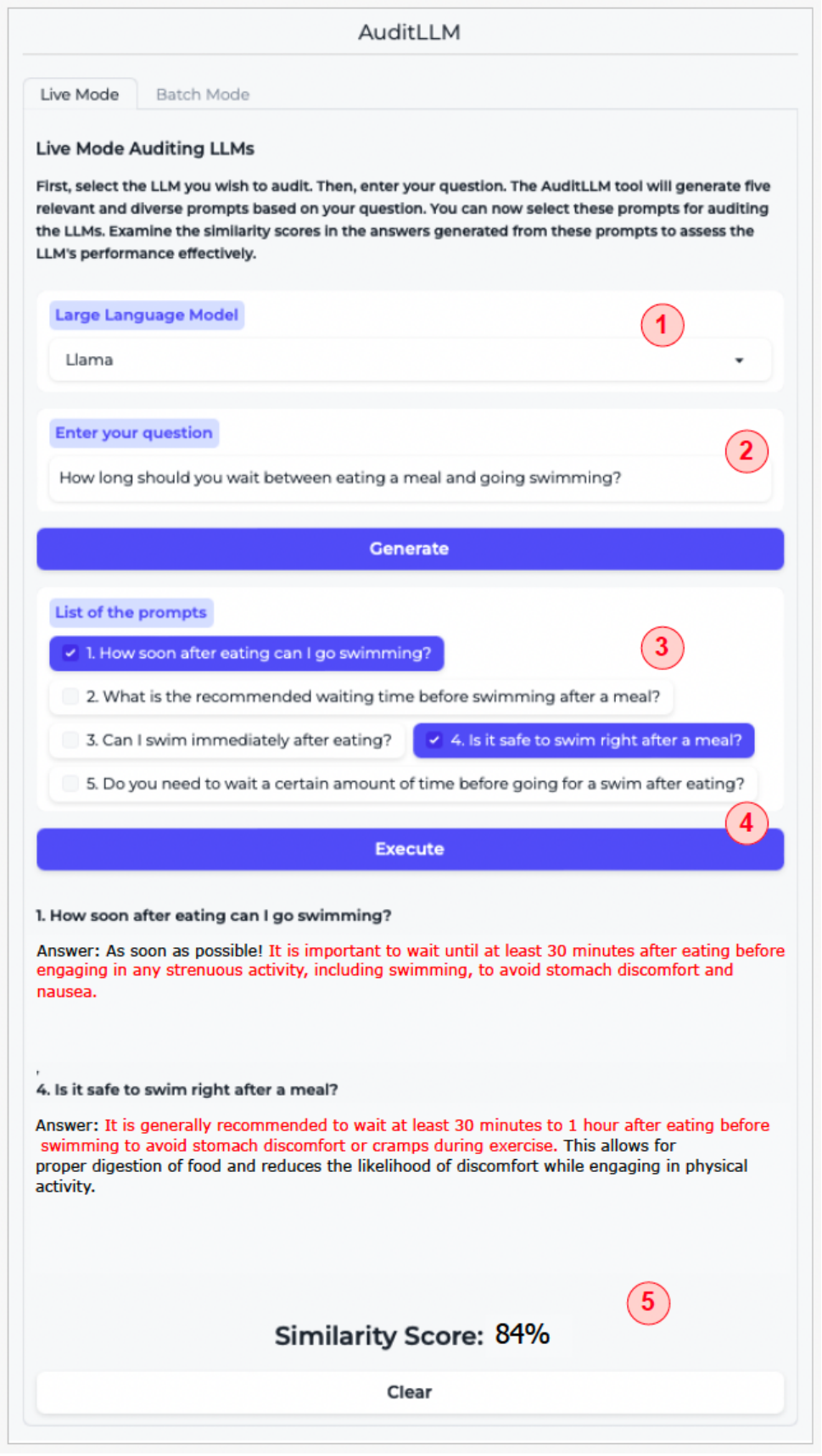}
    \caption{AuditLLM - Live Mode.}
    \vspace{-12pt}
    \label{fig:image1}
\end{figure}

The live mode allows for a real-time interrogation of LLMs through the submission of individual queries, providing immediate insights into the model's potential inconsistencies. The live mode comprises the following components, as illustrated in Figure~\ref{fig:image1}:

\begin{itemize}[leftmargin=0.2cm]
    \item \textbf{Step 1:}
        Users are provided with the capability to select from a predefined list of LLMs within the AuditLLM tool for auditing purposes. Currently available models include Llama 2-7B~\cite{Touvron2023LLAMA}, Falcon~\cite{penedo2023refinedweb}, Zephyr 7B~\cite{tunstall2023zephyr}, Vicuna~\cite{chiang2023vicuna}, and Alpaca~\cite{taori2023alpaca}. More models are being added through ongoing developments.
    \item \textbf{Step 2:}
        Users are required to input a query, which serves as the basis for auditing the LLM. 
    \item \textbf{Step 3:}
        Upon receiving user input, AuditLLM initiates the creation of customized probes aimed at evaluating the designated LLM. The prompt template for LLM1 is generated using the method outlined in~\cite{amirizaniani2024developing}, which utilizes human-in-the-loop validation and is based on criteria of relevance and diversity. We are using Mistral 7B LLM~\cite{jiang2023mistral} with temperature of 0.0 as our LLM1 to generate five probes that are not only closely related to the user's initial query but also cover a wide range of inquiries, enabling a comprehensive auditing.The prompt template employed is depicted in Figure~\ref{fig:template}.
    \item \textbf{Step 4:}
        Users are then afforded the opportunity to select from the generated probes, identifying those that most effectively facilitate the identification of inconsistencies within the audited LLM. This step is critical for narrowing down the focus of the audit to areas of potential concern or interest.
    \item \textbf{Step 5:} \label{sec:BERTScore}
        In the final step, the audited LLM (LLM2 in Figure\ref{fig:image1}) generate responses to the user-selected probes from the previous step. The semantic similarity of these responses is quantitatively determined by leveraging the BERTScore~\citep{zhang2019bertscore} via the sentence-transformer variant of MPNet (\texttt{all-mpnet-base-v2}) and cosine similarity, aiming to highlight sentences with semantic similarity. The culmination of this step is the presentation of a semantic similarity score, providing a metric for comparing responses and thus gauging the LLM consistency in processing and understanding queries.
\end{itemize}

Figure~\ref{fig:image1} showcases an application of this mode in the AuditLLM tool, following the steps previously outlined. AuditLLM highlights the semantic similarity between responses, applying a 60\% similarity threshold. Semantically similar sentences are highlighted with their similarity scores across texts, underscoring the consistent quality of the responses to these probes. This analysis of similarities across responses allows us to identify any inconsistencies in the model's outputs. A robust and reliable LLM should produce semantically consistent responses to similar questions, irrespective of their phrasing.

\begin{figure}[ht]
    \centering
    \includegraphics[width=0.40\textwidth]{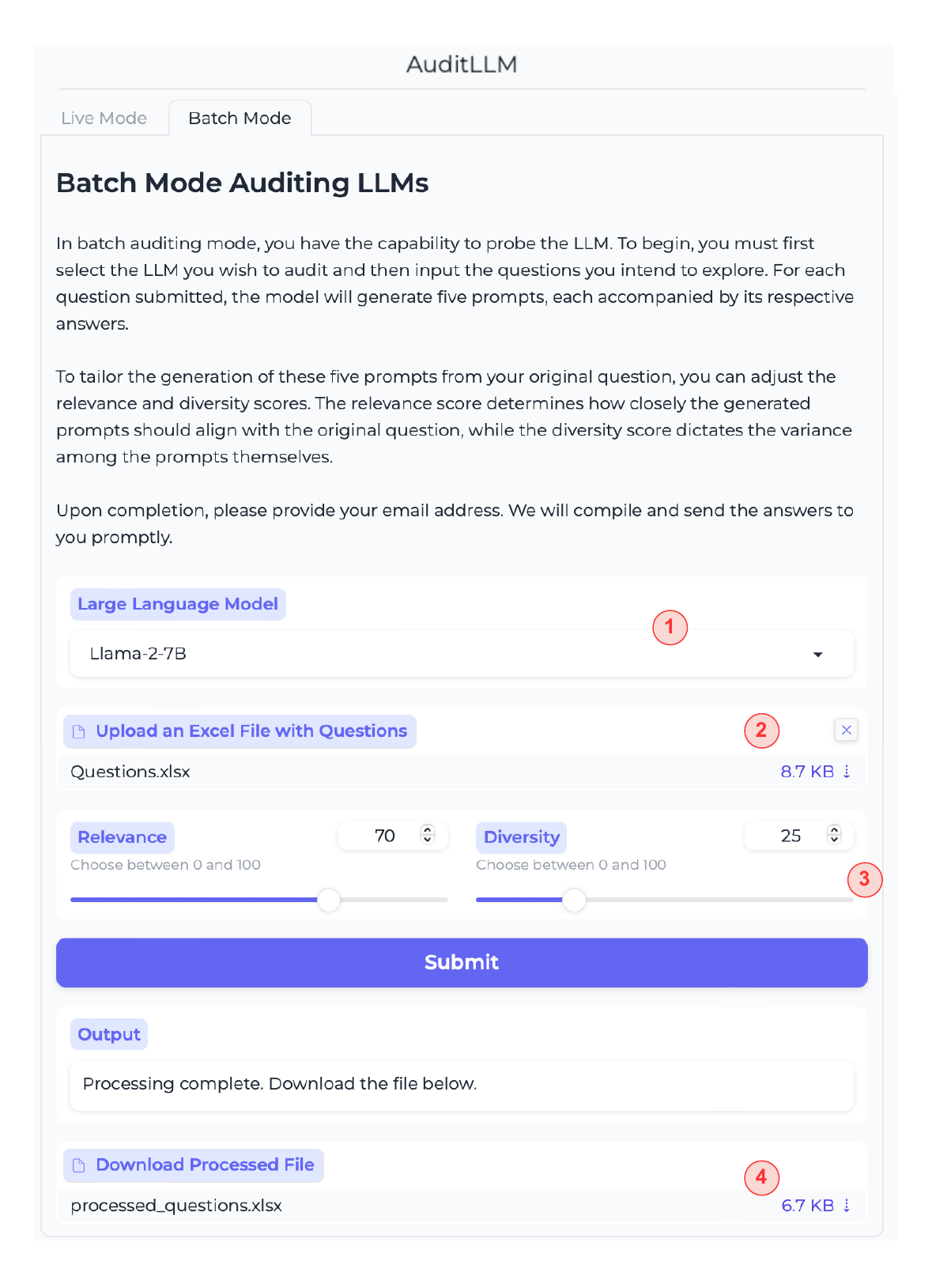}
    \caption{AuditLLM - Batch Mode.}
    \vspace{-12pt}
    \label{fig:batch}
    \vspace{-5pt}
\end{figure}

\subsection{Batch Mode}
The batch mode is designed for broader, more comprehensive assessments, allowing for the simultaneous submission of multiple questions for auditing the model's inconsistencies. The batch mode incorporates the following components, as delineated in Figure~\ref{fig:batch}:

\begin{itemize}[leftmargin=0.2cm]
    \item \textbf{Step 1:}
        The AuditLLM tool offers users the flexibility to choose from a curated selection of the above mentioned LLMs for the purpose of auditing. 
    \item \textbf{Step 2:}
        Users are requested to upload a set of questions in .xlsx format to serve as the basis for auditing the LLM.
    \item \textbf{Step 3:}
        In this step, we build upon existing research in the 'Question Generation' task, developing new probes from user questions with a focus on the criteria of relevance \citep{dugan-etal-2022-feasibility} and diversity \citep{raina2022multiplechoice}. This methodology requires that each probe closely aligns with the original user query and exhibits variety among them. Users are given the option to assign scores to each criterion, determining the extent to which the generated probes should exhibit relevance and diversity. These scores are then incorporated into the above mentioned prompt template, guiding the generation process to maintain query relevance and encompass diverse probes. For the probe generation, AuditLLM utilizes a Mistral 7B LLM set at a temperature of 0.0.
    \item \textbf{Step 4:}
        AuditLLM retrieves responses from the audited LLM via tailored probes, and assesses semantic similarity using the BERTScore, similar to the Section~\ref{sec:BERTScore} , via the sentence-transformer to identify closely related sentences. This includes calculating a semantic similarity score to quantitatively evaluate consistency. AuditLLM aggregates these findings, along with relevant details, and makes them available in an Excel format, ready for users to download.
\end{itemize}

\begin{figure}[ht]
    \centering
    \includegraphics[width=0.38\textwidth]{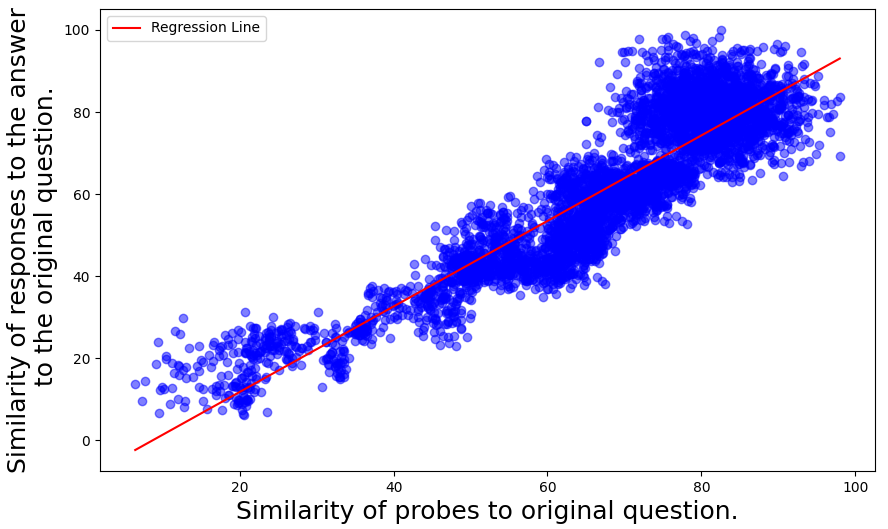}
    \caption{Prompt Similarity vs. Responses Similarity.}
    \vspace{-12pt}
    \label{fig:image2}
\end{figure}

To evaluate the effectiveness of AuditLLM, we conducted a batch mode run using questions from the TruthfulQA dataset \citep{lin-etal-2022-truthfulqa} as inputs. The results are displayed in Figure~\ref{fig:image2}. The x-axis shows the similarity of the five probes to the original question. The y-axis shows the similarity of the generated responses (by LLM2) to the answer for the original question as found in the dataset.
We expect that as the similarity among the probes goes up, so should the similarity among the responses. In a well-balanced output, the regression line should have a slope of 45$^{\circ}$. 
A higher slope indicates more similarity or consistency among the responses even when the probes are less similar (more diverse). A lower slope indicates greater divergence in responses for similar probes, indicating higher inconsistencies with the given LLM.
One could use the batch mode to run the same set of questions through different LLMs and compare their output distribution or regression lines to compare these LLMs for their consistencies and robustness.

\section{Conclusion}

AuditLLM provides a methodical auditing tool that examines the consistency of LLM responses to variably phrased queries. Its primary purpose is to evaluate a given language model by utilizing multiple probes generated from a single question. This methodology identifies discrepancies in the model's understanding or performance, including potential biases, hallucinations, and other related issues~\cite{amirizaniani2024developing}. Consequently, this process improves the reliability of these models in practical applications. With features like Live and Batch modes, AuditLLM offers valuable insights into the performance of LLMs through both real-time and extensive probing, making it an essential resource for both researchers and general users seeking to validate and improve the operational integrity of these complex systems.

AuditLLM, while innovative, does have its limitations. Specifically tailored for probing LLMs within the realm of textual data, it utilizes inconsistencies in responses as the primary mechanism for detecting potential issues. Despite incorporating six open-source LLMs recognized for their robust performance on the Huggingface leaderboard\footnote{~\url{https://huggingface.co/spaces/HuggingFaceH4/open_llm_leaderboard}},  it cannot audit the full spectrum of available LLMs. Additionally, given the substantial size of these models, loading them to generate responses can be time-intensive. Furthermore, variability in responses is observed when identical questions are administered in successive rounds, leading to disparate outcomes if a question is reiterated, as the LLMs yield unique answers. Moreover, AuditLLM’s reliance on the LangChain\footnote{~\url{https://www.langchain.com}} library requires that any modifications to LangChain prompt corresponding updates to AuditLLM to ensure continued functionality and compatibility. We are continuing to refine this tool by adding support for more LLMs, plugging in other ways to create multiple probes (e.g., using personas), and providing more methods for assessing inconsistencies.

\vspace{1em}
\noindent
{\bf ACKNOWLEDGEMENTS}

\noindent
This work was supported by the Bill and Melinda Gates Foundation.
\newpage

\bibliographystyle{ACM-Reference-Format}
\bibliography{Ref}

\end{document}